\documentclass{article}

\usepackage{microtype}
\usepackage{graphicx}
\usepackage{subfigure}
\usepackage{booktabs} % for professional tables
\usepackage{algorithmic}
\usepackage{graphicx}
\usepackage{textcomp}
\usepackage{xcolor}
\usepackage{enumitem}
\usepackage[numbers]{natbib}
\usepackage[ruled,vlined,algo2e]{algorithm2e}
\usepackage{float}

\usepackage{hyperref}

\newcommand{\projectname}{CATP}

\usepackage[accepted]{icml2024}

\usepackage{amsmath}
\usepackage{amssymb}
\usepackage{mathtools}
\usepackage{amsthm}

\usepackage[capitalize,noabbrev]{cleveref}

\theoremstyle{plain}

\theoremstyle{definition}

\theoremstyle{remark}

\usepackage[textsize=tiny]{todonotes}

\icmltitlerunning{CATP: Cross-Attention Token Pruning for Accuracy Preserved Multimodal
Model Inference}

\begin{document}

\twocolumn[
\icmltitle{CATP: Cross-Attention Token Pruning for Accuracy Preserved Multimodal Model Inference}



\icmlsetsymbol{equal}{*}
\begin{icmlauthorlist}
\icmlauthor{Ruqi Liao}{equal}
% \icmlaffiliation{Harvard University}
\icmlauthor{Chuqing Zhao}{equal}
\icmlauthor{Jin Li}{equal}
\icmlauthor{Weiqi Feng}{equal}
\icmlauthor{Yi Lyu}{}
\icmlauthor{Bingxian Chen}{}
\icmlauthor{Haochen Yang}{}
\end{icmlauthorlist}
\vspace{0.5em}
\centerline {Harvard University, University of Wisconsin-Madison, Stanford University}

% \icmlaffiliation{harvard}{Harvard University}
\icmlcorrespondingauthor{}{}

\icmlkeywords{Machine Learning, Model Pruning, Multi-modal Models}

\vskip 0.3in
]



\printAffiliationsAndNotice{\icmlEqualContribution} % otherwise use the standard text.

\begin{abstract}
In response to the rising interest in large multimodal models, we introduce Cross-Attention Token Pruning (\projectname{}), a precision-focused token pruning method. Our approach leverages cross-attention layers in multimodal models, exemplified by BLIP-2, to extract valuable information for token importance determination. \projectname{} employs a refined voting strategy across model heads and layers. In evaluations, \projectname{} achieves up to 12.1X higher accuracy compared to existing token pruning methods, addressing the trade-off between computational efficiency and model precision.
\end{abstract}

\section{Introduction}
\subsection{Background}
% \textbf{BLIP-2 Model.} 
The BLIP-2 is a state-of-the-art multi-modal model proposed by Li et al.\cite{li2023blip}. It enables a Large Language Models (LLMs) to understand images, and could be applied to various vision-language tasks such as image captioning, visual question answering (VQA) and chat-like conversation with an input of images. 

\begin{figure}[h]
    \centering
    \includegraphics[scale=0.5]{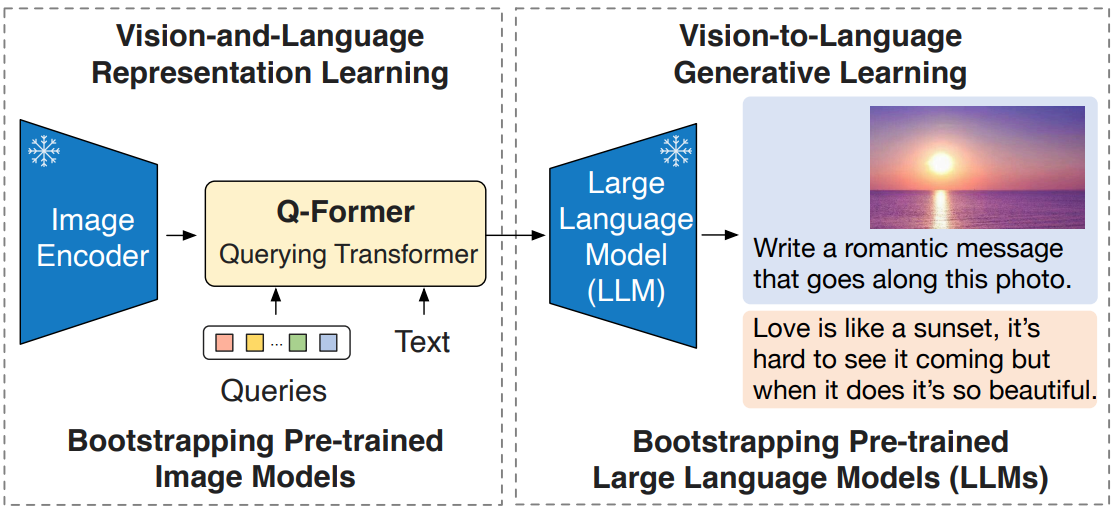}
    \caption{Overview of BLIP-2 Model's Framework\cite{li2023blip}}
    \label{fig:blip2-framework}
\end{figure}

As shown in Figure \ref{fig:blip2-framework}, BLIP-2 model consists of a frozen image encoder, a frozen LLM and a Querying Transformer (Q-Former) bridging the vision-language modality gap. Q-Former comprises two modules: an image transformer and a text transformer. These modules work to extract visual representation relevant to the corresponding text. 

\subsection{Challenges}
\textbf{LLM dominates BLIP-2 inference time.} In total, the BLIP-2 model has 3.1 billion parameters: 0.3 billion for Image Encoder, 0.1 billion for Q-Former and 2.7 billion for LLM Decoder. In other words, LLM Decoder accounts for over 87\% of total parameters, responsible for major computation cost of the BLIP-2 model during inference. 

\textbf{State-of-the-art pruning strategies reduce model accuracy.} Though state-of-the-art pruning strategies are effective in reducing computational costs for the LLM Decoder, they often result in a significant degradation of model accuracy when applied on BLIP-2. In our initial step, we evaluate two SOTA pruning methods--magnitude-based pruning and self-attention probability pruning algorithm \cite{wang2021spatten}. Both methods exhibit low accuracy in VQA task, which we will further discuss in Section VI. 

% \subsection{Goal}
\textbf{Accuracy Preserved Query-token Pruning.} Since the time-complexity and space-complexity of the attention mechanism in transformers are quadratic in the length of the input, the computational cost of LLM Deocoder can be reduced if the \texttt{seq\_len} dimension of the query token input is reduced. 

With these challenges in mind, we aim to design an end-to-end \textit{post-training pruning} pipeline that maintain decent accuracy for large-scale multi-modal models. 

\begin{figure*}[t]
    \centering
    \includegraphics[scale=0.5]{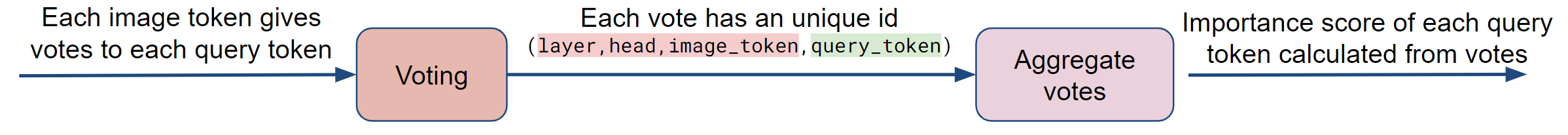}
    \caption{Voting procedure to decide less important query token that will be pruned.}
    \label{fig:vote}
\end{figure*}

\section{Methodology}
\begin{figure}[ht]
    \centering
    \includegraphics[scale=0.5]{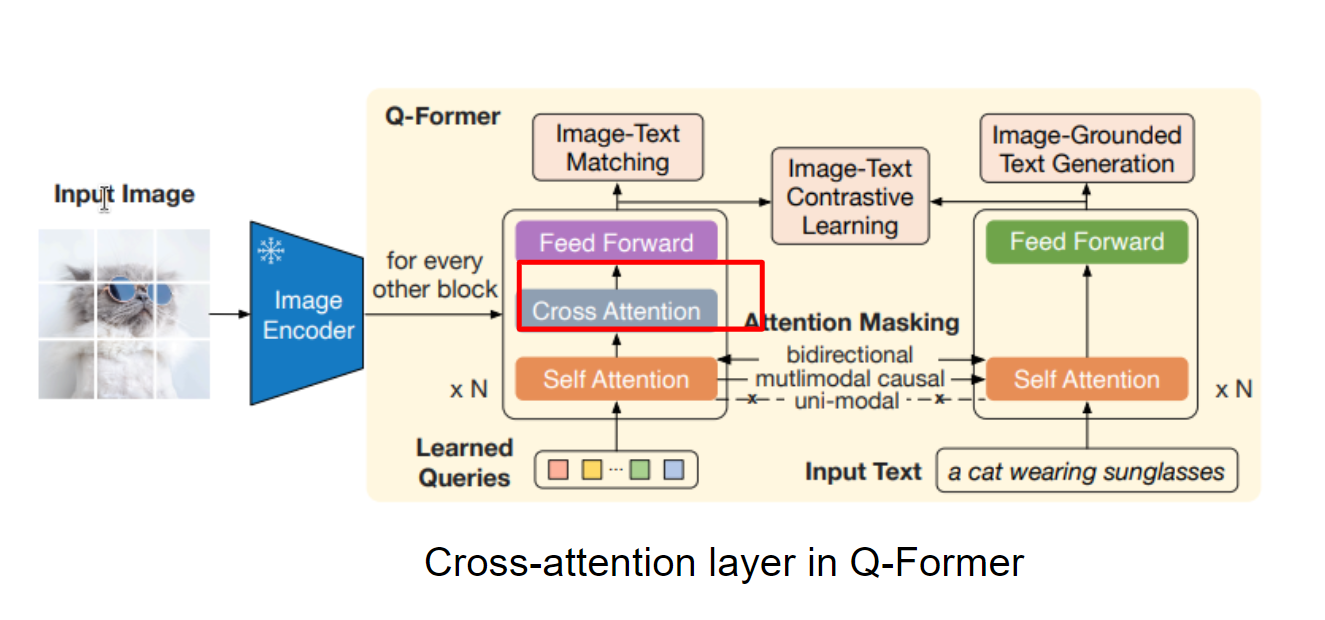}
    \caption{BLIP-2 model architecture of Q-Former\cite{li2023blip}. Cross-attention layers are inserted into Q-Former. This enables to incorporate information from both text and image inputs, and generate contextualized output. }
    %\textbf{(Right)} Tokens with darker colors in the cross-attention map have higher probabilities.}
    \label{fig:catp}
\end{figure}

We propose \textbf{C}ross-\textbf{A}ttention \textbf{T}oken \textbf{P}runing, a newly designed token-wise pruning method for BLIP-2. CATP suggests that the importance of each input query token of the LLM Decoder depends on its cross-attention probabilities with the input image tokens in the Q-Former. This is because the cross-attention probabilities indicate the relevancy between each image token and each query token and keeping the query tokens that are closely related to all image tokens from being pruned is necessary to preserving model accuracy. However, we cannot compute the importance of each query token by simply accumulating its cross-attention probabilities across all image tokens because they are computed using the softmax function and would sum up to 1. To address this issue, CATP additionally proposes a scheme that translate these probabilities into importance scores: for each query token, every image token votes different amount of points to it based on the cross-attention probability between them. As shown in Algorithm \ref{alg:1}, CATP takes the multi-head cross-attention layers inserted every other block in Q-Former (layer position marked in Figure \ref{fig:catp} and extracts cross-attention probability map from each head. For each map, every image token votes $L_{0}-n$ points to the query token with the $nth$ largest cross-attention probability where $L_{0}$ is the total number of query tokens. The importance score of each query token are computed by accumulating the points it received across layers and heads. The voting procedure is illustrated in Figure \ref{fig:vote}. We prune the query tokens with the top $L_{0}\times p$ lowest importance where p is the prune ratio. 

\begin{algorithm}[ht]
\small
\SetAlgoLined
\ $L_0$: number of query tokens \\
\ $L_1$: number of image patches \\
\ $h$: number of heads\\
\ Previous cumulative token importance score: $imp\_score \in L_0$\\
\ Current points assigned to token: $s \in  L_0$\\
%\ $query\_tokens \in\mathbb{R}^{B \times L_0 \times D}$ \\
\ $\text{Pruning\_ratio}: p$ \\
\ Cross-attention probability: $prob \in \mathbb{R}^{h \times L_0 \times L_1}$ \\
\For{$token_{id}$ = 0 to $L_{0}$}{
\For {$image_{id}$ = 0 to $L_{1}$} {
\For {$head_{id}$ = 0 to $h$} {

            \texttt{/* Voting Stage    */} \\
             $s[token_{id}]$ = \text{Vote}($prob[head_{id}][image_{id}][token_{id}]$) \\ 
              \texttt{/* Aggregate Votes   */} \\
              $imp\_score[token_{id}]$ += $s[token_{id}]$

}
}
}   
\newcommand{\TopK}[2]{\text{TopK}(#1, #2)}

\ $remained\_token\_id = \TopK{s}{L_0 \times (1 - p)}$
\caption{Cross-attention Pruning (one layer and one batch)}
\label{alg:1}
\end{algorithm}

%intellectual points
Although model compression has been a popular field for years, how to prune multimodal models is still under-explored. Many established pruning methods and rules that perform well on models with single modality cannot achieve acceptable results on multimodal models. Our project is able to provide a satisfactory solution that outperforms SOTA pruning strategies. By utilizing the cross-attention layers, CATP takes into account of all information from different modalities when computing the importance of query tokens and is able to make prune decisions that are more accurate than other pruning methods. In addition, CATP also proposes a new way of understanding the relationship between different modality inputs by applying voting strategies in analyzing cross-attention probabilities.

\section{Experiments}
In this section, we first describe the setups of our experiments including models, datasets and benchmarks. Then we show that \projectname{} performs better than the SOTA methods.
\subsection{Setups}
\subsubsection{Model and Dataset}
We evaluated \projectname{} on the \textbf{Blip-2 (blip2-opt-2.7b)} model and applied it to the \textbf{Visual Question Answering (VQA)} dataset with a sampling rate of 10\%. This subset was deemed adequate to assess the performance of our method. In our approach, we computed importance scores in two distinct ways to investigate their impact on model performance:
\begin{enumerate}
    \item For the first variant of \projectname{}, importance scores were computed utilizing cross-attention probabilities across all attention layers of QFormer.
    \item For the second variant, importance scores were derived exclusively from the first attention layer's cross-attention probabilities.
\end{enumerate}
\subsubsection{\textbf{Baselines}}
We compared our approach against two baselines for a comprehensive performance analysis:
\begin{enumerate}
    \item \textbf{L2-norm Baseline:} This method selects query tokens with the largest L2 norm under the assumption that tokens with higher magnitudes are more informative for the task.
    \item \textbf{Self-attention Baseline:} It selects query tokens with the highest self-attention scores as per the algorithm proposed in the SpAtten\cite{wang2021spatten} paper.
\end{enumerate}

In the subsequent subsections, we present a comparative analysis of \projectname{} against the established baselines.

\subsection{Main Results}

The results of applying \projectname{} to the VQA dataset are presented in Table \ref{tab:evaluation_results}. These results illustrate the performance of \projectname{} in comparison to the baselines and highlight the improvements achieved through the application of cross-attention based pruning.

\begin{table}[h]
\centering
\caption{Evaluation of query-token pruning on the 10\% VQA dataset}
\label{tab:evaluation_results}
\small
\begin{tabular}{lcc}
\hline
\textbf{Method} & \textbf{Pruning Ratio} & \textbf{Acc (\%)} \\
\hline
BLIP-2 model & No Pruning & 52.14 \\
L2-norm Baseline & Keep 1/2 & 9.50 \\
 & Keep 1/4 & 4.75 \\
 & Keep 1/8 & 8.91 \\
Self-attention Baseline & Keep 1/2 & 44.22 \\
 & Keep 1/4 & 7.29 \\
 & Keep 1/8 & 1.92 \\
\projectname{} (all layers/\textbf{first layer}) & Keep 1/2 & 41.14/\textbf{44.17} \\
 & Keep 1/4 & 31.15/\textbf{35.44} \\
 & Keep 1/8 & 23.27/\textbf{30.17} \\
\hline
\end{tabular}
\end{table}

\projectname{} demonstrates a performance improvement of up to 6.6X over the L2-norm baseline and up to 12.1X over the self-attention baseline. These improvements are particularly notable when pruning on the last layer's cross-attention probabilities. The difference between two variants of our methods is particularly interesting. In later section, we will study the factor of layer importance. 

\begin{figure}
    \centering
    \includegraphics[width=1\linewidth]{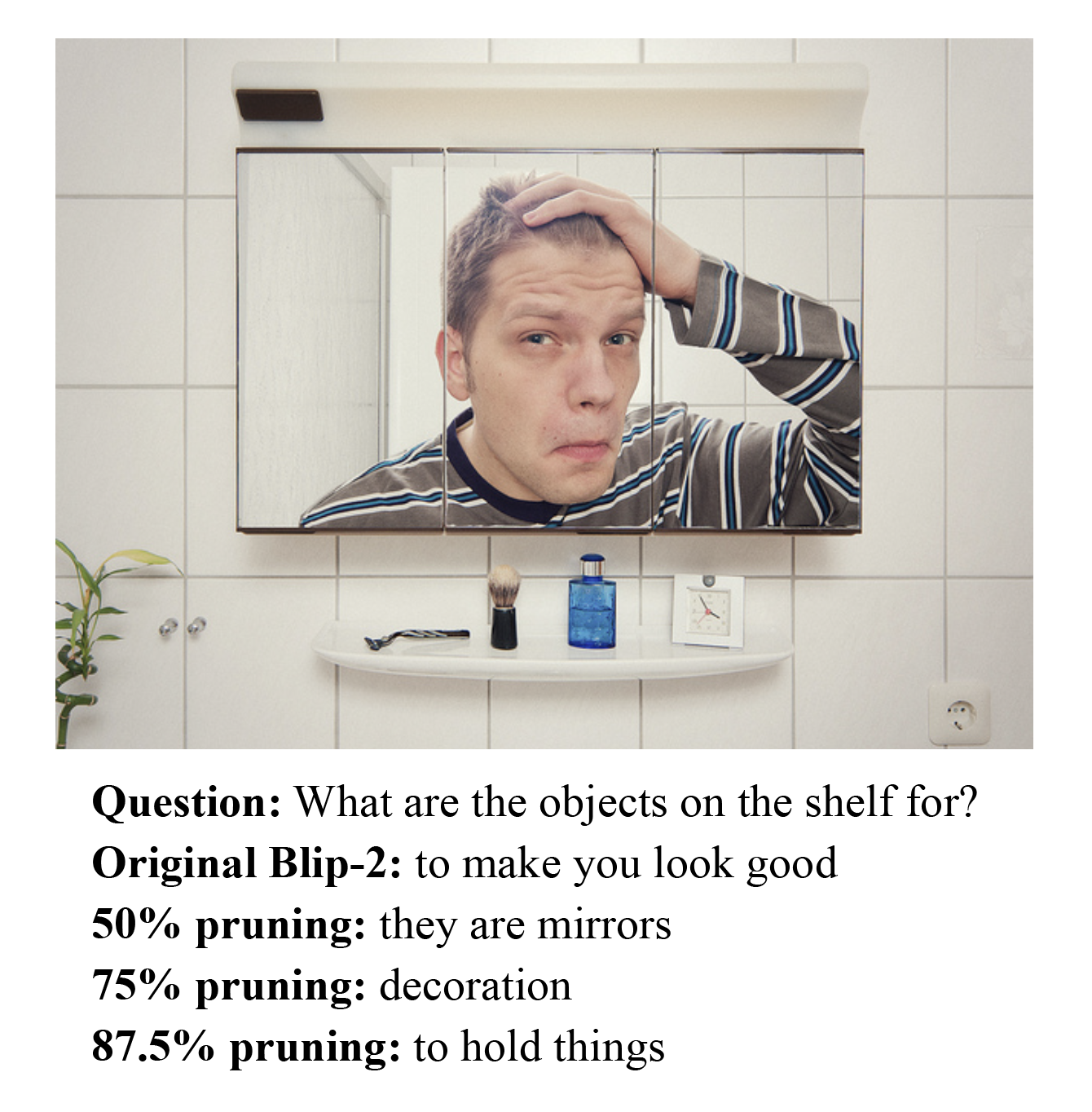}
    \caption{\centering A Visual-Question-Answering Example: comparing model output at different pruning levels}
    \label{fig:vqa-example}
\end{figure}

Figure \ref{fig:vqa-example} shows an example of the model output for the visual question answering task before and after the model being pruned at different level.
These results emphasize the potential of cross-attention mechanisms in enhancing model performance in query-token pruning tasks.

\subsection{Exploring Weighted Voting Further}
\subsubsection{voting weights of image tokens }
In an effort to further refine the query-token pruning process of \projectname{}, we have advanced our method to account for image-token importance. This enhancement is informed by the premise, as postulated by the SpAtten algorithm, that tokens with larger self-attention scores carry greater significance. Under this new scheme, image tokens are assigned varying weights derived from normalizing their self-attention scores, essentially allowing for a weighted voting mechanism in the pruning process.

The implementation of this advanced version utilizes the last attention layer of the visual encoder exclusively. The weighted algorithm is applied in conjunction with the first variant of our method, which leverages self-attention scores from all attention layers of QFormer. This approach is designed to harmonize the depth of attention insights across the model with the individual contribution of each image token, thereby enhancing the overall pruning strategy.

The results of incorporating image-token importance weighting are depicted in the table \ref{tab:image_token_importance}, which compares the performance of the standard cross-attention pruning (variant 1 in Table \ref{tab:evaluation_results}) against the performance when image patch importance weighting is applied.

\begin{table}[htbp]
\centering
\small
\caption{Evaluation of pruning with image-token weighted voting}
\label{tab:image_token_importance}
\begin{tabular}{lcc}
\hline
\textbf{Method} & \textbf{Pruning Ratio} & \textbf{Acc (\%)} \\
\hline
\projectname~(all layers)& Keep 1/2 & 41.14 \\
 & Keep 1/4 & 31.15 \\
 & Keep 1/8 & 23.27 \\
Pruning with weighted voting & Keep 1/2 & 41.90 \\
 & Keep 1/4 & 33.33 \\
 & Keep 1/8 & 29.30 \\
\hline
\end{tabular}
\end{table}

The data indicates that image-token importance weighting provides a notable improvement in model accuracy, particularly at more aggressive pruning ratios.

% Weiqi
\subsubsection{Layer importance} Further experiments illustrate that the information from cross-attention layers is not uniformly valuable for \projectname{}.

% \textbf{Experiment Setup:} 
We evaluate inference accuracy on 10\% of the VQA dataset when using a single cross-attention layer as the input for \projectname{}. Considering the BLIP-2 model comprises 6 cross-attention layers, we run \projectname{} for each layer and measure the accuracy at various pruning levels (retaining $\frac{1}{2}$, $\frac{1}{4}$, and $\frac{1}{8}$ of the tokens).

\begin{figure}[ht]
    \centering
    \includegraphics[scale=0.4]{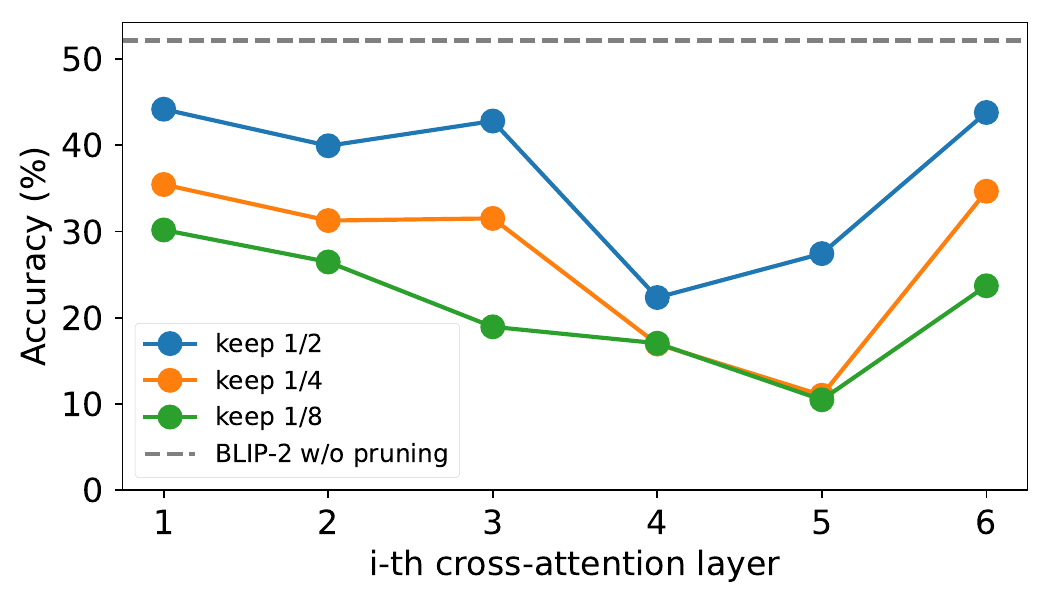}
    \caption{Inference accuracy across various cross-attention layers}
    \label{fig:layer_importance}
\end{figure}

% \textbf{Results:} 
Figure \ref{fig:layer_importance} displays the BLIP-2 inference accuracy following the execution of \projectname{} on each individual cross-attention layer. Our observation reveals that the initial and final cross-attention layers offer more valuable information to \projectname{}, resulting in up to 1.9X, 3.2X, and 2.9X higher inference accuracy compared to utilizing middle layers.

\section{Related works}
% Weiqi
% comparison with state of the art in the literature
\textbf{Large Multimodal Models.} There is a growing emphasis on large multimodal models capable of comprehending diverse types of data modalities, such as text, images, and videos. Examples of such models include Salesforce Research's BLIP-2 \cite{li2023blip}, OpenAI's GPT-4, and Google's recently introduced Gemini. Given that BLIP-2 is open-source on GitHub and has released its pretrained weights, \projectname{} opts to leverage the BLIP-2 model for evaluating the cross-attention pruning idea.

\textbf{Self-attention based token pruning.} SpAtten \cite{wang2021spatten} prunes tokens in NLP models based on self-attention probabilities, calculating token importance scores by aggregating attention probabilities across heads and layers. In contrast, \projectname{} focuses on query-token pruning for multimodal models (e.g., BLIP-2) and employs cross-attention to determine the importance of query-tokens.

\section{Conclusion}
We introduce \projectname{}, a cross-attention token pruning method aimed at maintaining the accuracy of multimodal model inference. Our approach begins by utilizing cross-attention probabilities to assess token importance. Additionally, we propose a novel voting strategy designed to aggregate cross-attention probabilities and estimate importance scores. In experiments, \projectname{} demonstrates an increase of up to 12.1x times in accuracy on the VQA dataset compared to state-of-the-art solutions, showcasing its effectiveness. Furthermore, our results indicate that incorporating image-token importance and layer-importance into the voting strategy has the potential to further enhance inference accuracy. 

\bibliographystyle{icml2024}
\bibliography{main}

\end{document}